\documentclass[10pt, conference, compsocconf]{IEEEtran} 
\IEEEoverridecommandlockouts
\usepackage{cite}
\usepackage{amsmath,amssymb,amsfonts}
\usepackage{algorithm} 
\usepackage{algorithmic}
\usepackage{graphicx}
\usepackage{textcomp}
\usepackage{xcolor}
\usepackage{booktabs}
\usepackage{caption}
\usepackage{array}
\usepackage{multirow}
\usepackage{subcaption}
\usepackage{tikz}
\usepackage{pgfplots}
\usepackage{url}

\pgfplotsset{compat=1.18}
\def\BibTeX{{\rm B\kern-.05em{\sc i\kern-.025em b}\kern-.08em
    T\kern-.1667em\lower.7ex\hbox{E}\kern-.125emX}}

\begin{document}

\title{Correlation to Causation: A Causal Deep Learning Framework for Arctic Sea Ice Prediction\\
}

\author{
    \IEEEauthorblockN{
        Emam Hossain\IEEEauthorrefmark{1}\IEEEauthorrefmark{2},
        Muhammad Hasan Ferdous\IEEEauthorrefmark{1}\IEEEauthorrefmark{2},
        Jianwu Wang\IEEEauthorrefmark{2}, 
        Aneesh Subramanian\IEEEauthorrefmark{3},
        Md Osman Gani\IEEEauthorrefmark{1}\IEEEauthorrefmark{2}
    }
    \IEEEauthorblockA{
        \IEEEauthorrefmark{1}Causal AI Lab, University of Maryland Baltimore County, USA \\
        \IEEEauthorrefmark{2}Department of Information Systems, University of Maryland Baltimore County, USA \\
        \IEEEauthorrefmark{3}Department of Atmospheric and Oceanic Sciences, University of Colorado Boulder, USA \\
        Email: \textit{emamh1@umbc.edu, h.ferdous@umbc.edu, jianwu@umbc.edu, aneeshcs@colorado.edu, mogani@umbc.edu}
    }
}

\maketitle

\begin{abstract}
Traditional machine learning and deep learning techniques rely on correlation-based learning, often failing to distinguish spurious associations from true causal relationships, which limits robustness, interpretability, and generalizability. To address these challenges, we propose a causality-driven deep learning framework that integrates Multivariate Granger Causality (MVGC) and PCMCI+ causal discovery algorithms with a hybrid deep learning architecture. Using 43 years (1979–2021) of daily and monthly Arctic Sea Ice Extent (SIE) and ocean-atmospheric datasets, our approach identifies causally significant factors, prioritizes features with direct influence, reduces feature overhead, and improves computational efficiency. Experiments demonstrate that integrating causal features enhances the deep learning model's predictive accuracy and interpretability across multiple lead times. Beyond SIE prediction, the proposed framework offers a scalable solution for dynamic, high-dimensional systems, advancing both theoretical understanding and practical applications in predictive modeling.
\end{abstract}


\section{Introduction}
Machine learning (ML) and deep learning (DL) have transformed predictive modeling across various domains, demonstrating remarkable performance in uncovering complex patterns from vast datasets. However, these models primarily rely on correlation-based learning, which presents inherent limitations. They often struggle to differentiate between spurious correlations and true causal relationships, leading to reduced robustness, interpretability, and generalizability \cite{pearl2018book}. For example, while ML models may capture statistical associations in a dataset, they often fail when deployed in new scenarios where causal mechanisms deviate from the observed data.

Causality addresses these challenges by uncovering the underlying cause-and-effect relationships that govern a system. Unlike correlation-based methods, causal discovery techniques such as Multivariate Granger Causality (MVGC)~\cite{barnett2014mvgc} and PCMCI+ \cite{runge2020discovering} identify direct and indirect factors of change, allowing models to focus on variables that genuinely influence outcomes. MVGC, an extension of traditional Granger Causality \cite{granger1969investigating}, is specifically designed for multivariate systems, providing computational efficiency and robustness for high-dimensional time series data. Incorporating causality into ML/DL models enhances predictive reliability, improves interpretability, and enables generalization to unseen environments.

\begin{figure}[!h]
\centering
\includegraphics[width=0.9\columnwidth]{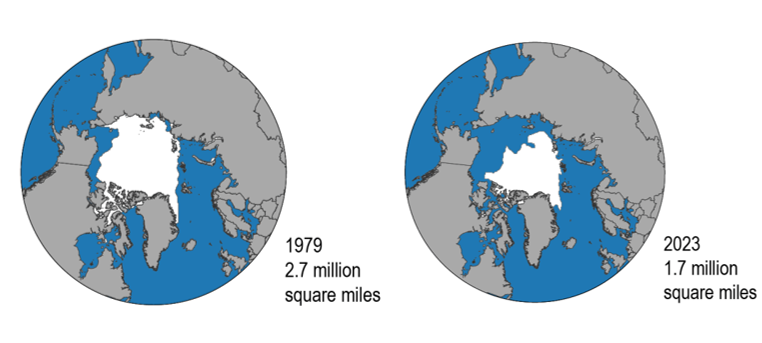}
\caption{Approximately 38\% decline in Arctic September sea ice, from 2.7 million square miles in 1979 to 1.7 million square miles in 2023 (Source: US Global Change Research Program).}
\label{fig:arctic_trends}
\end{figure}

The limitations of correlation-based approaches are especially articulated in dynamic, high-dimensional systems like climate forecasting. Arctic sea ice extent (SIE) prediction, which involves nonlinear interactions between oceanic and atmospheric variables, exemplifies these challenges. The rapid decline of Arctic sea ice (Figure~\ref{fig:arctic_trends}) has profound implications for global weather patterns, ecosystems, and human activities.
However, conventional forecasting methods and correlation-driven ML/DL models fail to provide reliable predictions for long lead times, as they struggle to capture the complex causal interactions in Arctic climate systems \cite{andersson2021seasonal}.

To address these gaps, this study introduces a causality-driven framework that integrates causal discovery techniques with deep learning models for Arctic SIE prediction. Using 43 years (1979–2021) of ocean-atmospheric datasets, we identify causally significant features using MVGC and PCMCI+. These features serve as input to a hybrid deep learning architecture comprising Gated Recurrent Units (GRUs) and Long Short-Term Memory (LSTM) networks. This architecture enables the model to prioritize factors directly influencing sea ice dynamics, reduce feature overhead, and improve computational efficiency. This causality-guided approach not only enhances the robustness and interpretability of predictions but also simplifies the model by focusing on a subset of causal variables, thereby addressing the limitations of traditional correlation-based methods.

The key contributions of this work are threefold. (1) We employ MVGC and PCMCI+ to identify the causal relationships between Arctic SIE and ocean-atmospheric variables, providing an interpretable understanding of the key factors of change. (2) We design and implement a hybrid GRU-LSTM model that integrates the identified causal features to improve short- and long-term predictive accuracy. (3) We empirically evaluate the proposed framework using metrics such as RMSE, MAE, and \(R^2\), demonstrating significant improvements in forecast reliability for lead times of up to six months. By addressing the limitations of correlation-based models, this work bridges the gap between causality and predictive modeling.

\section{Related Works}
This section reviews key advancements in Arctic sea ice prediction and causality-driven predictive modeling.

\subsection{ML/DL in Arctic Sea Ice Prediction}
ML models have been widely applied to Arctic sea ice prediction due to their ability to process large datasets and model nonlinear dependencies. Zhu et al. \cite{zhu2024stdnet} proposed the Spatio-Temporal Decomposition Network (STDNet) for improved Arctic sea ice concentration forecasts, while Driscoll et al. \cite{driscoll2024data} developed a data-driven emulator for melt pond prediction by integrating physical models with ML. Similarly, Koo and Rahnemoonfar \cite{koo2024hierarchical} introduced a hierarchical convolutional neural network (CNN) that combines multiple ice-related parameters to enhance forecast precision. DL models have further advanced predictive capabilities by capturing complex spatial and temporal dynamics. Luo et al. \cite{xu2024sifm} employed a foundational DL model for multi-granularity sea ice forecasting, while Ren et al. \cite{ren2024sicnet} utilized transformer-based architectures to integrate sea ice thickness data into seasonal predictions. Kim et al. \cite{kim2025long} demonstrated the potential of U-Net-based models by incorporating climatic variables such as surface temperature and radiation into long-term forecasts.  Liu et al. \cite{liu2024physics} introduced a physics-informed DL model for predicting sea ice concentration and velocity.
While these models have advanced the field, they remain limited by their reliance on statistical correlations, which can lead to overfitting and reduced interpretability in dynamic systems like Arctic climate forecasting \cite{dunmireSGL2025}.

\subsection{Causality-Driven Predictive Modeling}
Recognizing the limitations of correlation-driven approaches, recent research has incorporated causality into ML/DL frameworks.
For instance, Oliveira et al. \cite{oliveira2024causality} integrated causal insights into financial time series forecasting, demonstrating improved generalization and interpretability. Li et al. \cite{li2024advancing} applied Granger Causality to Arctic sea ice prediction, highlighting its utility in uncovering dynamic environmental relationships. 
Building on these advancements, this work integrates MVGC and PCMCI+ with a hybrid GRU-LSTM architecture to predict Arctic SIE. By prioritizing causally significant ocean-atmospheric variables, the framework reduces feature redundancy and improves long- and short-term forecast accuracy. Empirical results validate its effectiveness for complex, high-dimensional systems like the Arctic climate, demonstrating the critical role of causality in robust and interpretable predictive modeling.

\section{Background}
This section introduces the foundations of causal discovery algorithms like MVGC and PCMCI+, and recurrent neural networks (RNNs), particularly LSTM and GRU. 

\subsection{Causal Discovery}
\begin{figure}[h!]
\centering
\includegraphics[width=0.9\columnwidth]{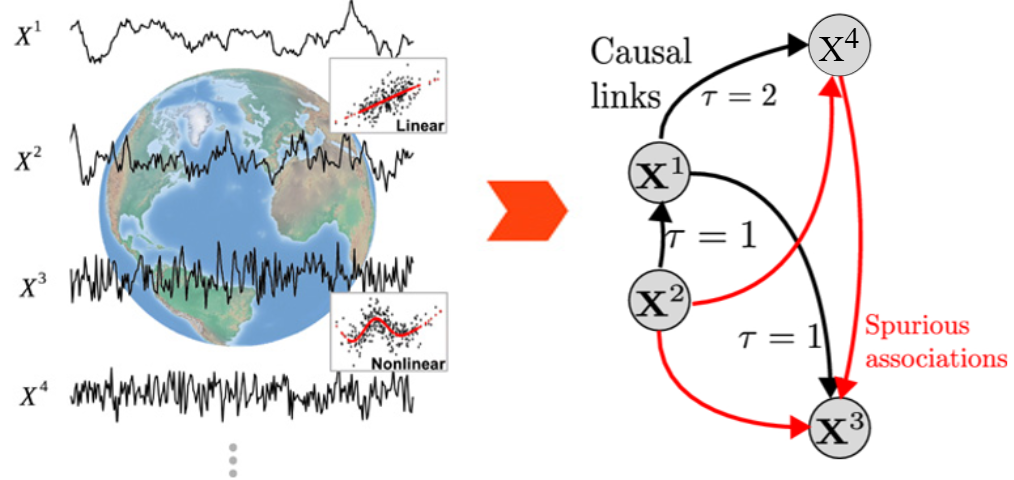}
\caption{Illustration of how causal discovery algorithms identify lagged causal relationships from time series data. $\tau$ represents the timelag of the causal links.}
\label{fig:causal_discovery_process}
\end{figure}

Causal discovery identifies cause-and-effect relationships within time series data, providing insights beyond simple statistical associations \cite{hasansurvey}. From the observational data, it constructs a causal graph (Figure~\ref{fig:causal_discovery_process}) that shows the causal connecting among the variables, and their associated time lags \cite{ferdous2023cdans}. This enhances robustness, interpretability, and generalizability in predictive modeling, particularly in systems with nonlinear interactions.

\textbf{Multivariate Granger Causality (MVGC)} \cite{barnett2014mvgc}, an extension of Granger Causality (GC) proposed by Granger \cite{granger1969investigating}, evaluates whether lagged values of multiple time series collectively improve the prediction of another variable. Unlike traditional GC, MVGC is designed for multivariate systems, making it more suitable for high-dimensional datasets. Its computational efficiency and ability to handle interactions among multiple variables make it particularly effective in identifying causal relationships in complex environmental systems such as the Arctic.

\textbf{PCMCI+} \cite{runge2020discovering}, on the other hand, extends the PC (Peter-Clark) algorithm with momentary conditional independence (MCI) testing. It is robust to autocorrelated and high-dimensional datasets, effectively disentangling direct causal links from indirect ones. PCMCI+ excels at identifying critical factors in environmental datasets dominated by spurious correlations \cite{hossain2024incorporating}.


\begin{figure*}[!h]
\centering
\includegraphics[width=1.8\columnwidth]{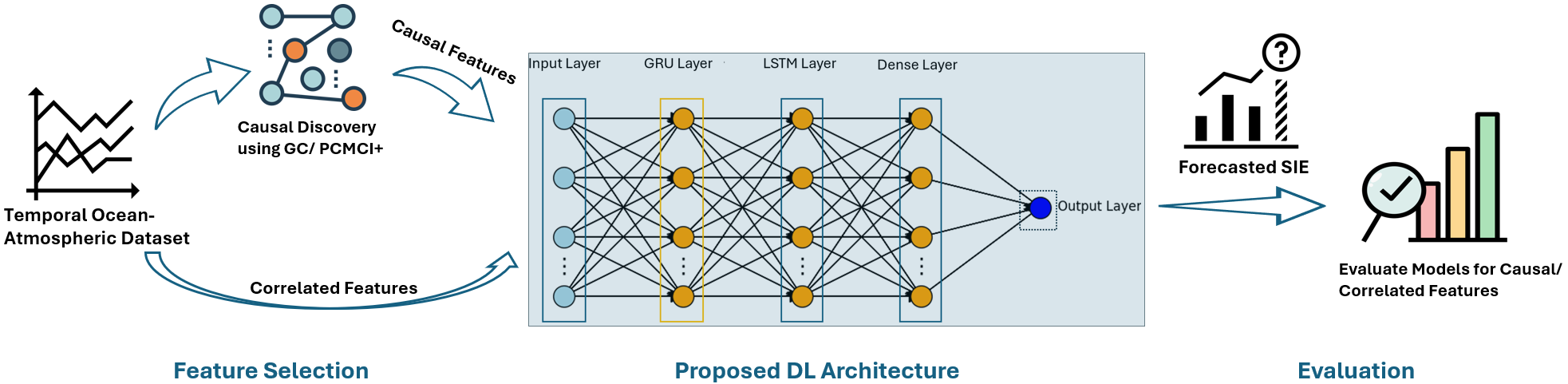}
\caption{Overview of the proposed causal deep learning framework for Arctic Sea Ice Extent (SIE) prediction. Daily and monthly temporal ocean-atmospheric datasets are analyzed with MVGC and PCMCI+ to identify causal features. These features are integrated into a GRU-LSTM architecture for SIE forecasting and performance is evaluated on causal vs. correlated features.}
\label{fig:methodology}
\end{figure*}


\subsection{Recurrent Neural Networks (RNNs)}
RNNs, including Gated Recurrent Units (GRU) and Long Short-Term Memory (LSTM) networks, are widely used for time series forecasting due to their ability to model temporal dependencies. These architectures address challenges like the vanishing gradient problem, enabling effective learning of both short- and long-term patterns.

\textbf{Gated Recurrent Unit (GRU)}~\cite{cho2014learning} simplifies RNN architectures with fewer parameters by combining update and reset gates. This makes GRUs computationally efficient while maintaining predictive accuracy. \textbf{Long Short-Term Memory (LSTM)} networks \cite{hochreiter1997long} employ memory cells and gating mechanisms to capture long-term dependencies. LSTMs are widely applied in time series forecasting for environmental, and climate datasets \cite{liu2024physics}. Combining GRU and LSTM leverages the computational efficiency of GRU and the long-term memory capabilities of LSTM. Recent studies demonstrate that hybrid GRU-LSTM models outperform standalone architectures \cite{hossain2020novel}, making them ideal for integrating causally identified features into predictive models.

\section{Methodology}
This section presents our proposed causality-driven framework for forecasting the Arctic Sea Ice Extent. The framework consists of four key components: (a) data collection \& preprocessing, (b) causal feature identification, (c) designing causal deep learning model, and (d) model training \& evaluation. The overall architecture is illustrated in Figure~\ref{fig:methodology}.

\subsection{Data Collection and Preprocessing}
This study leverages atmospheric and oceanic datasets alongside sea ice extent (SIE) data to analyze long-term trends and seasonal variations in Arctic climate dynamics \cite{ali2021sea}. The datasets are obtained from two primary sources: atmospheric and oceanic variables are derived from the ERA-5 global reanalysis product, while SIE values are calculated from sea ice concentrations based on the Nimbus-7 SSMR and DMSP SSM/I-SSMIS passive microwave data \cite{cavalieri1996sea} provided by the National Snow and Ice Data Center (NSIDC).

To ensure a comprehensive representation of Arctic climate dynamics, the study develops two distinct time series datasets. The first series consists of monthly gridded data, spatially averaged over the Arctic region north of 25° N using an area-weighted method, covering the years 1978 to 2021. The second series includes daily gridded data for the same spatial region, spanning 1979 to 2018, which facilitates the analysis of short-term variations and causal dependencies. The dataset encompasses 10 ocean-atmospheric variables and SIE values, as detailed in Table~\ref{tab:variables_summary}. To maintain consistency and reduce noise, standard preprocessing steps are applied, including normalization, imputation of missing values, and temporal aggregation. These steps ensure that the data is well-prepared for causal discovery, enabling the identification of critical factors influencing Arctic SIE dynamics.

\begin{table}[h!]
\centering
\caption{Summary of Daily \& Monthly Sea Ice Datasets.}
\label{tab:variables_summary}
\begin{tabular}{@{}lll@{}}
\toprule
\textbf{Variable}          & \textbf{Range}           & \textbf{Unit}       \\ \midrule
Surface Pressure           & [400, 1100]             & hPa                 \\
Wind Velocity              & [0, 40]                 & m/s                 \\
Specific Humidity          & [0, 0.1]                & kg/kg               \\
Air Temperature            & [200, 350]              & K                   \\
Shortwave Radiation        & [0, 1500]               & W/m\textsuperscript{2} \\
Longwave Radiation         & [0, 700]                & W/m\textsuperscript{2} \\
Rainfall                   & [0, 800]                & mm/day              \\
Snowfall                   & [0, 200]                & mm/day              \\
Sea Surface Temperature    & [200, 350]              & K                   \\
Sea Surface Salinity       & [0, 50]                 & psu                 \\
Sea Ice Extent             & [3.34, 16.63]           & Million             \\ \bottomrule
\end{tabular}
\end{table}

\subsection{Causal Feature Identification}
Causal feature identification plays a critical role in improving the interpretability and predictive accuracy of Arctic SIE forecasts. Building upon the previously introduced MVGC and PCMCI+ algorithms, we applied these methods to identify key causal variables of Arctic sea ice dynamics. For both daily and monthly datasets, \textbf{MVGC} identified all variables except \textit{Sea Surface Temperature (SST)} as causal features. This result underscores the broad influence of atmospheric and oceanic variables on Arctic sea ice.

\begin{figure}[h!]
\centering
\begin{subfigure}{\linewidth}
    \centering
    \includegraphics[width=0.85\linewidth]{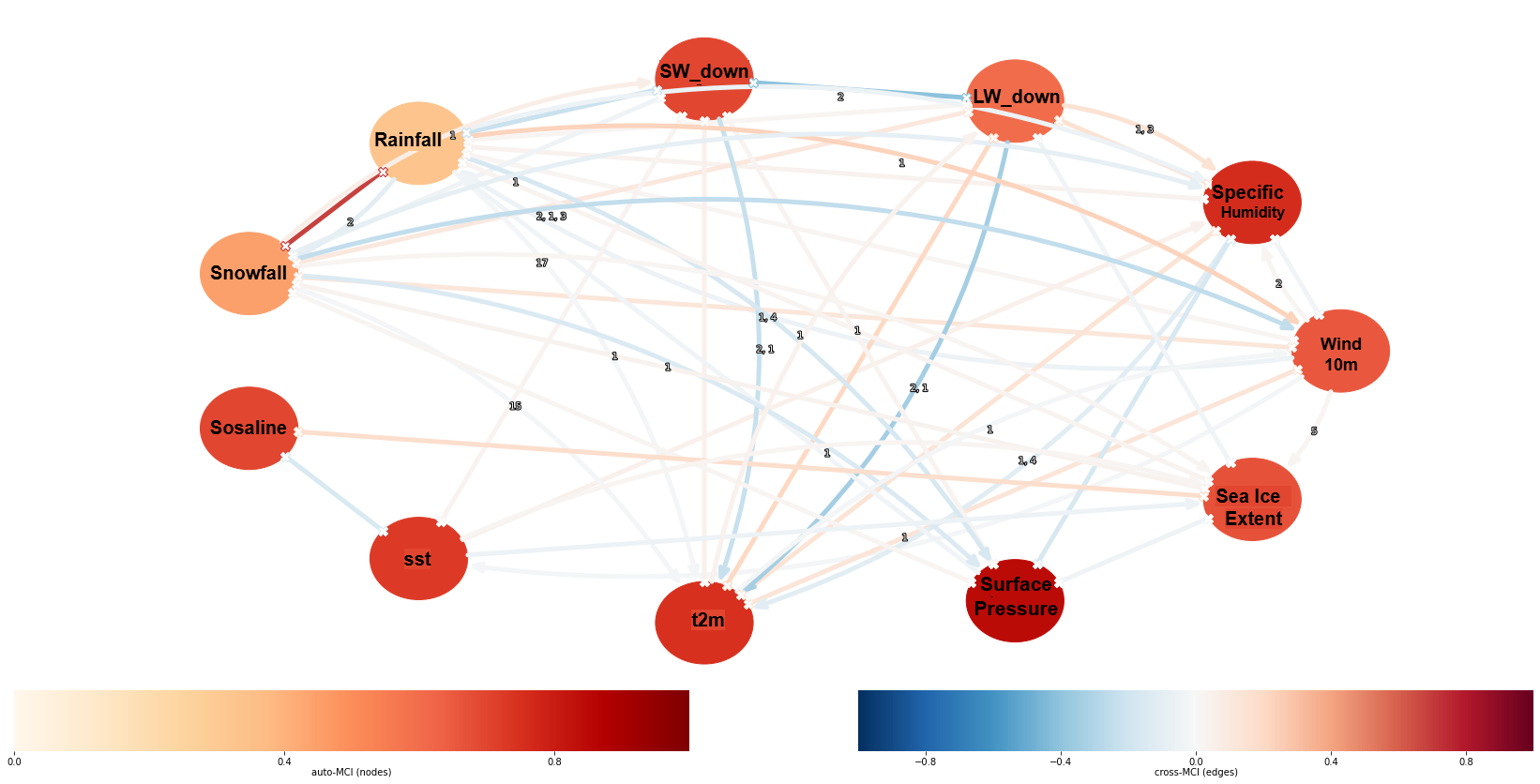}
    \caption{Causal graph of PCMCI+ for daily data.}
    \label{fig:causal_graph_daily}
\end{subfigure}
\vfill
\begin{subfigure}{\linewidth}
    \centering
    \includegraphics[width=0.85\linewidth]{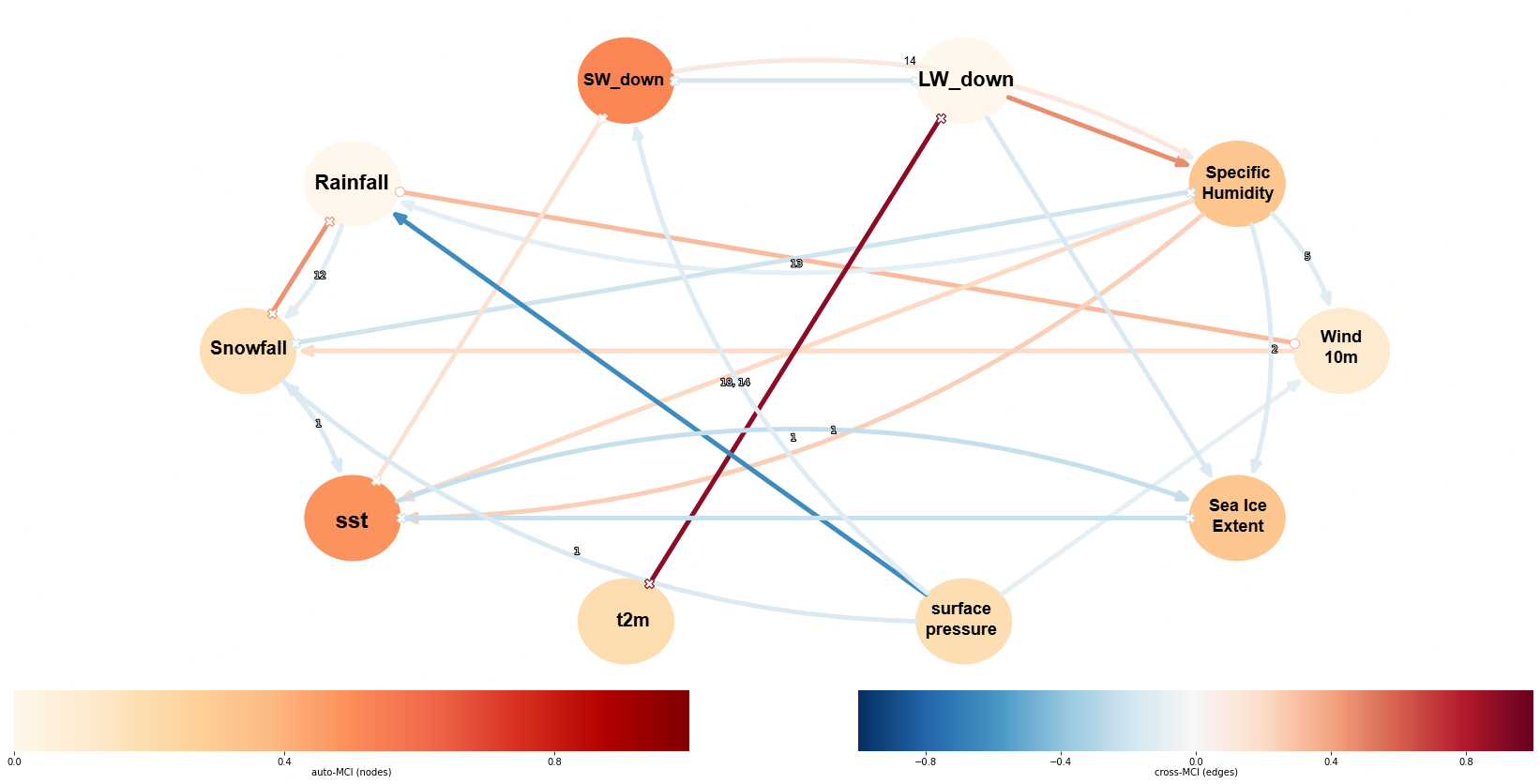}
    \caption{Causal graph of PCMCI+ for monthly data.}
    \label{fig:causal_graph_monthly}
\end{subfigure}
\caption{Causal graphs generated by PCMCI+ for (a) daily and (b) monthly datasets, illustrating the causal relationships between ocean-atmospheric variables and SIE.}
\label{fig:causal_graphs}
\end{figure}

\textbf{PCMCI+}, known for its robustness in handling high-dimensional and autocorrelated time series data, provided a more refined identification of causal features. For the daily dataset, PCMCI+ highlighted \textit{longwave radiation, snowfall, sea surface salinity (SSS), surface pressure}, and \textit{SIE} itself as the primary causal factors. For the monthly dataset, the identified causal features were \textit{longwave radiation, SST}, and \textit{SIE}. These results suggest temporal and spatial differences in the causal relationships influencing SIE dynamics across daily and monthly timescales. Figure~\ref{fig:causal_graphs} shows the causal graphs generated by PCMCI+ for daily and monthly datasets, highlighting the direct causal influences of key variables on Arctic SIE. The identified features guided the selection of input variables for the GRU-LSTM model, ensuring that the model leveraged causally significant information for prediction.

\subsection{Designing Causal Deep Learning Model}
To leverage the identified causal features, we developed a hybrid deep learning architecture combining Gated Recurrent Units (GRUs) and Long Short-Term Memory (LSTM) networks. This hybrid design effectively captures both short-term and long-term dependencies in time series data. The input time series is constructed with a lookback window of 21 timesteps, corresponding to the maximum time lag we considered for the causal discovery algorithms. The architecture comprises an input layer with 21 neurons for encoding past timesteps, a GRU layer with 64 neurons and 20\% dropout for short-term pattern extraction, an LSTM layer with 128 neurons and 20\% dropout for long-term dependencies, and a dense layer with 64 neurons for feature integration. Finally, the output layer with a single neuron predicts SIE across 1–6 month lead times (Figure \ref{fig:methodology}).

\begin{algorithm}[!h]
\caption{Causal Deep Learning Framework for Arctic Sea Ice Forecasting}
\begin{algorithmic}[1]
\label{alg:arctic_sie}
\STATE \textbf{Input:} Time series data $\mathcal{D} = \{X_t, Y_t\}$, where $X_t$ represents ocean-atmospheric variables and $Y_t$ is SIE. Maximum lag $\tau = 21$, causal discovery algorithms MVGC and PCMCI+.
\STATE \textbf{Output:} Forecasted SIE $\hat{Y}_{t+h}$ for lead times $h \in \{1, \dots, 6\}$.

\STATE \textbf{Step 1: Preprocessing}
\STATE Extract and preprocess daily and monthly datasets $\mathcal{D}_{\text{daily}}$ and $\mathcal{D}_{\text{monthly}}$ from ERA-5 and NSIDC.
\STATE Normalize features, impute missing values, and generate lagged variables up to $\tau$.

\STATE \textbf{Step 2: Causal Feature Identification}
\STATE Apply MVGC to identify causal variables $C_{\text{GC}} \subseteq X_t$.
\STATE Apply PCMCI+ to identify causal variables $C_{\text{PCMCI+}} \subseteq X_t$ for both daily and monthly datasets.

\STATE \textbf{Step 3: Model Training}
\STATE Define the GRU-LSTM architecture:
\begin{itemize}
    \item \textbf{Input Layer:} features from $C \in \{X_t, C_{\text{GC}}, C_{\text{PCMCI+}}\}$.
    \item \textbf{GRU Layer:} $64$ neurons with dropout rate $p=0.2$.
    \item \textbf{LSTM Layer:} $128$ neurons with dropout rate $p=0.2$.
    \item \textbf{Dense Layer:} $64$ neurons.
    \item \textbf{Output Layer:} Single neuron for predicting $\hat{Y}_{t+h}$.
\end{itemize}

\STATE Train models:
\begin{itemize}
    \item $\mathbf{DL_{\text{vanilla}}}$ using $X_t$.
    \item $\mathbf{DL_{\text{GC}}}$ using $C_{\text{GC}}$.
    \item $\mathbf{DL_{\text{PCMCI+}}}$ using $C_{\text{PCMCI+}}$.
    \item $\mathbf{DL_{\text{DPCMCI+}}}$ using $C_{\text{PCMCI+}}^{\text{daily}}$ for monthly predictions.
\end{itemize}

\STATE Train with \textit{Adam} optimizer, MSE loss function, batch size $64$, $100$ epochs, and early stopping to prevent overfitting.

\STATE \textbf{Step 4: Evaluation}
\STATE Evaluate all models on test data $\mathcal{D}_{\text{test}}$ using RMSE, MAE, and $R^2$ metrics for lead times $h \in \{1, \dots, 6\}$.
\end{algorithmic}
\end{algorithm}

\subsection{Model Training and Evaluation}
The GRU-LSTM models were trained using data up to 2013, with 10\% of the training data utilized for validation. The remaining data (2014–2018 for daily interval data, and 2014–2021 for monthly interval data) was used for testing. Separate models were trained for daily and monthly datasets to evaluate the impact of causal feature integration. For both cases, three primary model types were considered: \textbf{$DL_{\text{vanilla}}$}, trained on all 10 ocean-atmospheric variables; \textbf{$DL_{\text{GC}}$}, trained on features identified by Multivariate Granger Causality; and \textbf{$DL_{\text{PCMCI+}}$}, trained on features identified by PCMCI+. Additionally, for monthly datasets, a fourth model, \textbf{$DL_{\text{DPCMCI+}}$}, was included, utilizing PCMCI+ features derived from daily data but adapted for monthly predictions. Table~\ref{tab:model_inputs} shows the input features considered for each model. This technique allows for a comprehensive evaluation of how causal features enhance the predictive performance of the deep learning model.

\begin{table*}[!h]
\centering
\caption{Input features used for the trained models.}
\label{tab:model_inputs}
\resizebox{2\columnwidth}{!}{%
\begin{tabular}{@{}lll@{}}
\toprule
\textbf{Model} & \textbf{Datasets Used} & \textbf{Input Features} \\ \midrule
$DL_{\text{vanilla}}$  & Daily / Monthly & All 10 ocean-atmospheric variables \\
$DL_{\text{GC}}$  & Daily / Monthly & \textit{Surface Pressure, Wind Velocity, Specific Humidity, Air Temperature, Shortwave Radiation, Longwave Radiation, Rainfall, Snowfall, SSS, SIE} \\
$DL_{\text{PCMCI+}}$  & Daily & \textit{Surface Pressure, Longwave Radiation, Snowfall, SSS, SIE} \\
$DL_{\text{PCMCI+}}$  & Monthly & \textit{Longwave Radiation, SST, SIE} \\
$DL_{\text{DPCMCI+}}$  & Monthly & \textit{Surface Pressure, Longwave Radiation, Snowfall, SSS, SIE} \\ \bottomrule
\end{tabular}%
}
\end{table*}

The training process employed the \textit{Adam} optimizer with a mean squared error (MSE) loss function. Models were trained using a batch size of 64 over 100 epochs, with early stopping applied to avoid overfitting. Model performance was evaluated using three key metrics: Root Mean Squared Error (\textbf{RMSE}), quantifying the magnitude of prediction errors; Mean Absolute Error (\textbf{MAE}), measuring the average absolute deviation of predictions from the observed values; and the Coefficient of Determination (\(\mathbf{R^2}\)), assessing the proportion of variance in the data that the model explains. The datasets and codes are publicly available on GitHub\footnote{\url{https://github.com/ehfahad/Causal-DL-for-Arctic-SIE-Prediction}}. The complete workflow for the proposed causal deep learning technique for sea ice forecasting is outlined in Algorithm~\ref{alg:arctic_sie}.

\section{Results and Discussion}
\begin{figure}[h!]
\centering
\includegraphics[width=0.95\columnwidth]{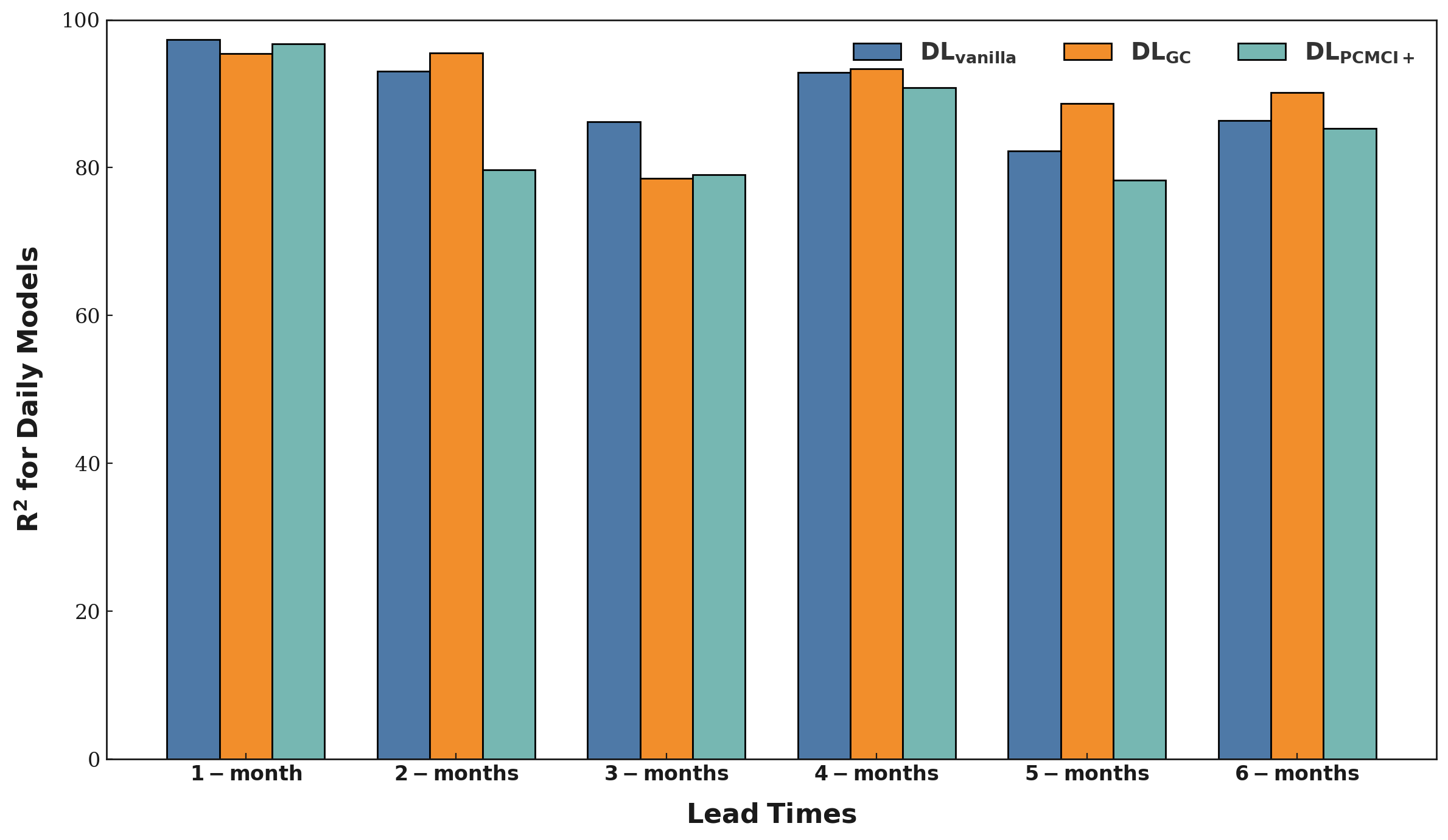}
\caption{\(R^2\) values for daily models across the lead times.}
\label{fig:R2_daily}
\end{figure}

This section evaluates the performance of the proposed causality-driven framework for Arctic Sea Ice Extent (SIE) prediction, showing how integrating causal features with deep learning techniques improves accuracy and interpretability. Table~\ref{tab:daily_metrics_full} presents the RMSE and MAE values for daily models trained on all features (\textbf{$DL_{\text{vanilla}}$}), Multivariate Granger Causality features (\textbf{$DL_{\text{GC}}$}), and PCMCI+ features (\textbf{$DL_{\text{PCMCI+}}$}). The corresponding \(R^2\) values are visualized in Figure~\ref{fig:R2_daily}. For daily models, $DL_{\text{vanilla}}$ achieves the best performance at 1-month lead times, as reflected in the lowest RMSE and highest \(R^2\). However, as the lead time increases, $DL_{\text{GC}}$ demonstrates superior performance at intermediate horizons, particularly for 2-, 4-, and 5-month forecasts. This highlights the capability of MVGC to capture long-term dependencies effectively. On the other hand, $DL_{\text{PCMCI+}}$ excels in short-term predictions, with the best MAE values for 1-month and 3-month lead times, emphasizing the utility of PCMCI+ in identifying immediate causal influences.

\begin{table}[h!]
\centering
\caption{Error metrics for daily models.}
\label{tab:daily_metrics_full}
\begin{tabular}{@{}ccccc@{}}
\toprule
\textbf{Lead Time} & \textbf{Metric} & $DL_{\text{vanilla}}$ & $DL_{\text{GC}}$ & $DL_{\text{PCMCI+}}$ \\ \midrule
1-month  & RMSE (\%) & \textbf{7.777} & 8.017  & 8.043  \\
         & MAE (\%)  & 4.856  & 5.556  & \textbf{4.625} \\ \midrule
2-months & RMSE (\%) & 14.303 & \textbf{11.365} & 21.663 \\
         & MAE (\%)  & 7.003  & \textbf{5.593}  & 9.441  \\ \midrule
3-months & RMSE (\%) & \textbf{18.200} & 24.381 & 22.465 \\
         & MAE (\%)  & 7.719  & 8.883  & \textbf{8.524} \\ \midrule
4-months & RMSE (\%) & 13.708 & \textbf{11.274} & 13.779 \\
         & MAE (\%)  & 7.659  & \textbf{7.091}  & 8.163  \\ \midrule
5-months & RMSE (\%) & 20.340 & \textbf{17.586} & 20.822 \\
         & MAE (\%)  & 9.207  & \textbf{7.683}  & 10.294 \\ \midrule
6-months & RMSE (\%) & \textbf{15.342} & 15.038 & 17.573 \\
         & MAE (\%)  & 9.083  & \textbf{7.210}  & 8.979  \\ \bottomrule
\end{tabular}
\end{table}

\begin{figure}[h!]
\centering
\includegraphics[width=0.95\columnwidth]{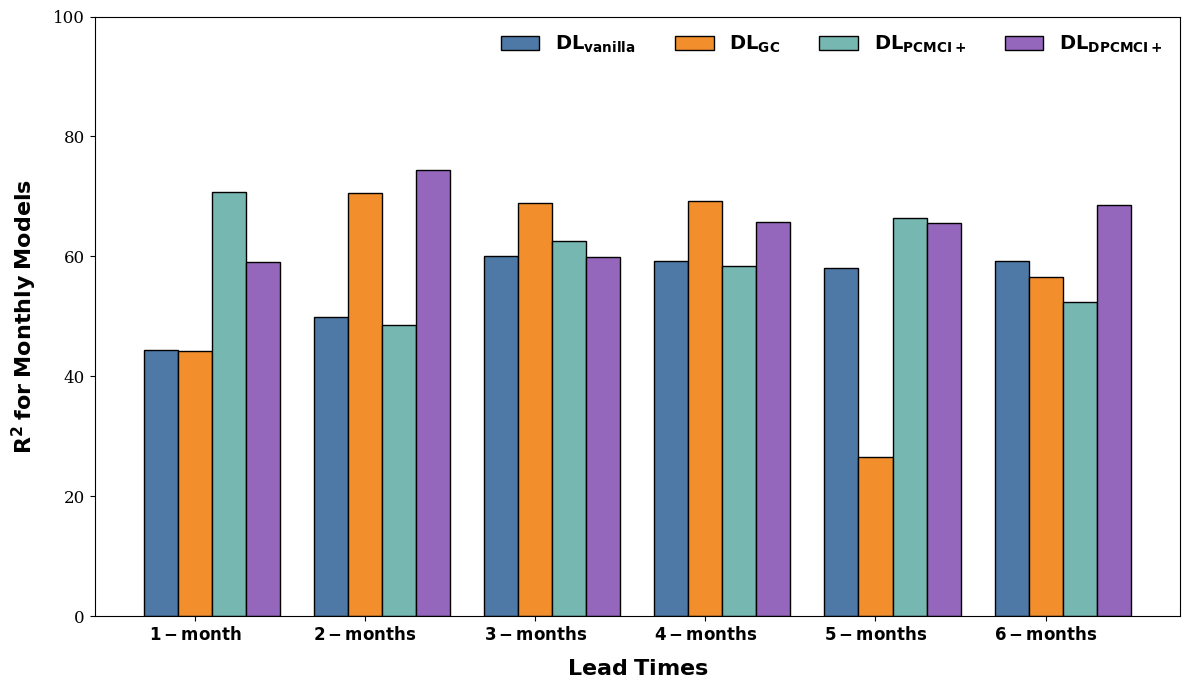}
\caption{\(R^2\) values for monthly models across the lead times.}
\label{fig:R2_monthly}
\end{figure}

For monthly models, overall the results demonstrate higher errors (Table~\ref{tab:monthly_metrics_full}) and lower \(R^2\) (Figure~\ref{fig:R2_monthly}) values compared to daily models. This performance disparity is attributed to the use of monthly averaged data, which smoothens temporal variability and limits the model’s ability to capture fine-grained dynamics. Among the monthly models, $DL_{\text{PCMCI+}}$ achieves the best RMSE and \(R^2\) for 1-month lead times, confirming the utility of PCMCI+ features for short-term predictions. At intermediate horizons (e.g., 2 months), $DL_{\text{DPCMCI+}}$ outperforms others by adapting causal features from daily data for monthly forecasting. This feature transfer preserves short-term causal relationships, enhancing long-term stability. The trend continues at 5–6 months, where $DL_{\text{DPCMCI+}}$ consistently achieves the best accuracy, highlighting the benefit of finer temporal features in extended forecasts.

\begin{table}[h!]
\centering
\caption{Error metrics for monthly models.}
\label{tab:monthly_metrics_full}
\resizebox{\columnwidth}{!}{%
\begin{tabular}{@{}cccccc@{}}
\toprule
\textbf{Lead Time} & \textbf{Metric} & $DL_{\text{vanilla}}$ & $DL_{\text{GC}}$ & $DL_{\text{PCMCI+}}$ & $DL_{\text{DPCMCI+}}$ \\ \midrule
1-month  & RMSE (\%) & 30.556 & 31.188 & \textbf{21.608} & 24.863 \\
         & MAE (\%)  & 16.169 & 15.839 & 16.884 & \textbf{15.839} \\ \midrule
2-months & RMSE (\%) & 27.081 & \textbf{23.451} & 26.040 & 19.851 \\
         & MAE (\%)  & 20.723 & 15.834 & 16.032 & \textbf{16.032} \\ \midrule
3-months & RMSE (\%) & 24.769 & \textbf{21.052} & 24.120 & 25.653 \\
         & MAE (\%)  & 20.156 & 18.397 & \textbf{18.783} & 19.676 \\ \midrule
4-months & RMSE (\%) & 23.007 & 22.675 & 25.133 & \textbf{22.105} \\
         & MAE (\%)  & 17.170 & \textbf{16.606} & 19.736 & 18.432 \\ \midrule
5-months & RMSE (\%) & 27.462 & 32.750 & \textbf{24.000} & 21.805 \\
         & MAE (\%)  & 19.522 & 18.324 & \textbf{17.835} & 18.949 \\ \midrule
6-months & RMSE (\%) & 27.815 & 27.247 & 26.971 & \textbf{21.883} \\
         & MAE (\%)  & 16.328 & 23.522 & 20.094 & \textbf{16.648} \\ \bottomrule
\end{tabular}%
}
\end{table}

Since no single model consistently outperforms the others, model selection should depend on the forecasting objective. For short-term predictions (1-month lead time), $DL_{\text{PCMCI+}}$ is preferable. At intermediate lead times, $DL_{\text{GC}}$ provides better generalization, while for extended forecasts (5-6 months), $DL_{\text{DPCMCI+}}$ benefits from daily-scale causal feature transfer. A hybrid approach that combines different models across lead times may further enhance performance. While these models show significant promise, challenges remain. Both PCMCI+ and MVGC are computationally intensive for large causal graphs with numerous environmental factors and are sensitive to hyperparameter settings, such as lag selection, which can impact scalability and stability. Optimizing these methods through reduced computational complexity \cite{hossain2024time} and adaptive lag selection could further enhance their practicality for operational Arctic sea ice forecasting.

\section{Conclusion}
This work introduces a causality-driven framework for forecasting the Arctic Sea Ice Extent (SIE), bridging the gap between correlation-based machine learning models and causality-focused approaches. By leveraging Multivariate Granger Causality (MVGC) and PCMCI+, we identify key causal factors among ocean-atmospheric variables, ensuring that the models prioritize features with a direct influence on SIE dynamics. This shift from correlation to causation addresses the limitations of traditional ML/DL models, reducing feature overhead, simplifying model complexity, and improving computational efficiency by excluding irrelevant variables. The hybrid GRU-LSTM architecture demonstrates significant improvements in predictive performance when trained on causality-identified features compared to models trained on correlated ones. Results show that causal feature integration enhances predictive accuracy across multiple lead times (1-6 months), as evidenced by higher \(R^2\), reduced RMSE, and lower MAE. 

This study highlights the potential of integrating causality into deep learning, offering a pathway to more interpretable, robust, and efficient prediction in dynamic and high-dimensional systems like the Arctic climate. Beyond Arctic sea ice prediction, the proposed framework can be adapted to other domains where causality is essential for understanding complex temporal dynamics. Future research will focus on optimizing the computational efficiency of causal discovery algorithms, adaptive lag selection, and extending the framework to address spatial heterogeneity and multi-scale interactions.

\section*{Acknowledgment}
This work is supported by iHARP: NSF HDR Institute for Harnessing Data and Model Revolution in the Polar Regions (Award\# 2118285). The views expressed in this work do not necessarily reflect the policies of the NSF, and endorsement by the Federal Government should not be inferred.

\bibliographystyle{IEEEtran} 
\bibliography{References}

\vspace{12pt}

\end{document}